\documentclass[letterpaper, 10 pt, conference]{ieeeconf}  % Comment this line out if you need a4paper

\IEEEoverridecommandlockouts                              % This command is only needed if 
\usepackage{multicol}
\usepackage[utf8]{inputenc} % allow utf-8 input
\usepackage[T1]{fontenc}    % use 8-bit T1 fonts
\usepackage{hyperref}       % hyperlinks
\usepackage{url}            % simple URL typesetting
\usepackage{booktabs}       % professional-quality tables
\usepackage{amsfonts}       % blackboard math symbols
\usepackage{nicefrac}       % compact symbols for 1/2, etc.
\usepackage{microtype}      % microtypography
\usepackage{epigraph}
\usepackage{graphicx}
\usepackage{amsmath}
\usepackage{array}
\usepackage{hyperref}
\usepackage[binary-units=true]{siunitx}
\usepackage{subcaption}
\usepackage{bm}
\usepackage{mathtools}
\usepackage{cuted}
\usepackage{etoolbox}
\usepackage{balance}
\usepackage{multirow}
\usepackage[skip=10pt,font=small]{caption}

\usepackage{wrapfig}

\newcommand\mypara[1]{\vspace{3pt}\noindent\textbf{#1.}}

\usepackage{pgfplots}
\pgfplotsset{compat=newest}
\usetikzlibrary{plotmarks}
\usetikzlibrary{babel}
\usetikzlibrary{arrows.meta}
\usepgfplotslibrary{patchplots}
\usepackage{grffile}
\usepackage{amsmath}
\usepackage{interval}

\usepgfplotslibrary{groupplots}
\usepgfplotslibrary{dateplot}
\title{\LARGE \bf
Learning a Single Near-hover Position Controller\\for Vastly Different Quadcopters
}

\author{Dingqi Zhang$^1$, Antonio Loquercio$^2$, Xiangyu Wu$^1$, Ashish Kumar$^2$, Jitendra Malik$^2$, and Mark W. Mueller$^1$% <-this % stops a space
\thanks{The authors are with the $^1$High Performance Robotics Lab, Dept. of Mechanical Engineering, and the $^2$Dept. of Electrical Engineering and Computer Science, University of California Berkeley, \{dingqi, loquercio, wuxiangyu, ashish\_kumar, malik, mwm\}@berkeley.edu} 
}

\begin{document}

\maketitle
\maketitle
\thispagestyle{empty}
\pagestyle{empty}

% \todo{formatting to fit 6+n}

% \todo{•	How do you initialize the history to estimate z in the real world. Specifically, what are the initial history to generate z, or how do you generate z for the first iteration?}

% \todo{I think similar to [1] you could do a PCA analysis on the embedding space to see how the mass or other parameters affect the embedding. }

% \todo{ Why did you use a 3-layer 1-D cnn for the adaption module, rather then a more sequential based model (which seems to fit well for the time-based nature of the data), for example a LSTM or Transformer.  }

% \todo{Minor Correction (Typos, etc.) Sec II, Paragraph 2: Define the intrinsics vector before using the term in passing.}

% \todo{The main criticisms involve the need of narrowing down the main contributions as well as the comparison to earlier work. In the event that these two criticisms are tackled, the paper might be suitable for publication.}

%%%%%%%%%%%%%%%%%%%%%%%%%%%%%%%%%%%%%%%%%%%%%%%%%%%%%%%%%%%%%%%%%%%%%%%%%%%%%%%%
\begin{abstract}
This paper proposes an adaptive near-hover position controller for quadcopters, which can be deployed to quadcopters of very different mass, size and motor constants, and also shows rapid adaptation to unknown disturbances during runtime. The core algorithmic idea is to learn a single policy that can adapt online at test time not only to the disturbances applied to the drone, but also to the robot dynamics and hardware in the same framework.
We achieve this by training a neural network to estimate a latent representation of the robot and environment parameters, which is used to condition the behaviour of the controller, also represented as a neural network. We train both networks exclusively in simulation with the goal of flying the quadcopters to goal positions and avoiding crashes to the ground. We directly deploy the same controller trained in the simulation without any modifications on two quadcopters in the real world with differences in mass, size, motors, and propellers with mass differing by 4.5 times. In addition, we show rapid adaptation to sudden and large disturbances up to one-third of the mass of the quadcopters. We perform an extensive evaluation in both simulation and the physical world, where we outperform a state-of-the-art learning-based adaptive controller and a traditional PID controller specifically tuned to each platform individually. Video results can be found at~\url{https://youtu.be/U-c-LbTfvoA}.
% Video results can be found \todo{at~\url{https://youtu.be/3yQrDML5aWs}}. %We will release our code and trained model upon acceptance.

\end{abstract}

% \makeatletter
% \g@addto@macro\@maketitle{
% \centering

% \vspace{-0.1in}
% \bigskip}
% \makeatother
% \textcolor{red}{We never use the second person ("you") in a paper. A paper isn't addressed at anyone, it is simply a report.}

%%%%%%%%%%%%%%%%%%%%%%%%%%%%%%%%%%%%%%%%%%%%%%%%%%%%%%%%%%%%%%%%%%%%%%%%%%%%%%%%
%===============================================================================

\section{Introduction}

\begin{figure}
    \includegraphics[width=\columnwidth]{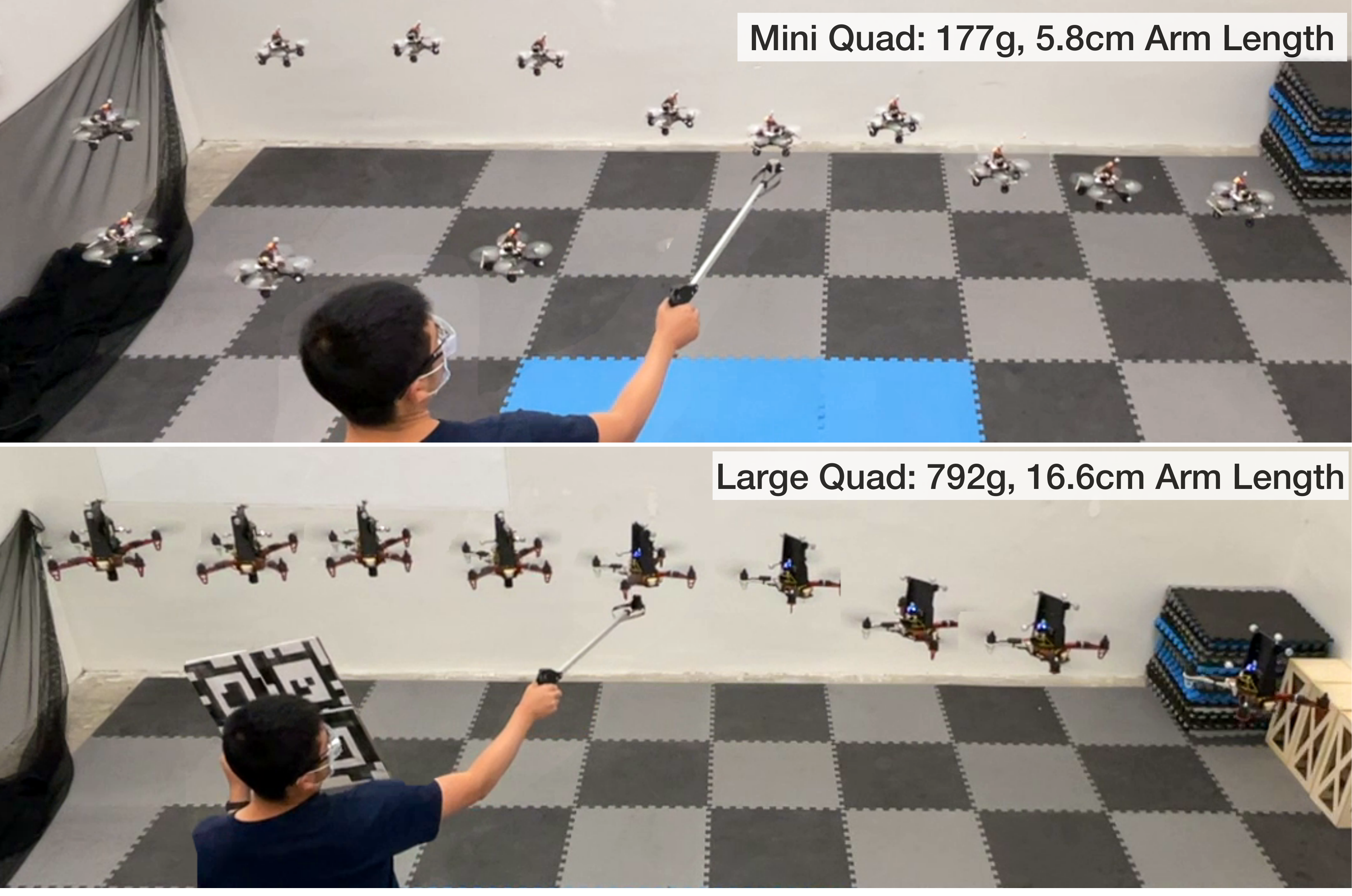}
\captionof{figure}{Demonstration of our adaptive controller on two quadrotors with widely varying mass, arm length, and motor constants for the task of tracking a straight and circular trajectory. In the middle of the trajectory, we add a payload unknown to the drone of approximately 30\% of the robot's mass. Our end-to-end controller is able to quickly estimate and react to the disturbance. We use a single control policy across different drones and tasks, which is deployed without any vehicle-specific modifications.}
\label{fig:teaser}
\end{figure}

A classic model-based controller for quadcopters requires explicit estimates of the vehicle's model, including inertia properties (mass and mass moment of inertia), motor constants, and other parameters.
Errors in estimating such parameters directly translate to errors in the execution of the controller commands.
In addition, once the parameters are estimated, the controller typically requires specific tuning in-flight to achieve the desired behaviour.
%
% If the estimation and tuning is done right, the final performance will be very good.
%
% But what looked like a simple task required significant engineering effort.
%
A modification to the vehicle, e.g., replacing the motors with ones that have different properties, would require repeating the above steps in estimation and tuning.
Such significant engineering efforts could be eased with a single controller without specialized tuning. However, this problem raises several challenges since quadcopters are unstable systems. 
Large variations in quadcopter design have prompted the design of specialized controllers for different quadcopters, as opposed to one single controller. 
Rejection to disturbances or model uncertainties requires non-trivial design changes to the specialized controllers, which add more layers of complexity to the design work.
In this work, we show a single near-hover position controller aiming to control a variety of quadcopters differing in mass, size, motor, propeller, etc. with mass and planar area size notably differing by at least 10 times. 
In addition, our controller can rapidly adapt to unknown in-flight disturbances in the mass and inertia changes of the quadcopters.
Our work builds on three main insights.
The first is that relevant dynamics parameters are observable from the history of states and actions.
This has long been known in the community: several works have shown how to estimate parameters online with nonlinear Kalman filters~\cite{svacha2020imu,wuest2019online} or with neural networks~\cite{forgione2021continuous}.
The second insight is that, at test time, we do not need an estimate of the parameters in some “ground truth” sense.
What matters is that the estimate leads to the “right” action, which our end-to-end training procedure optimizes for.
This simplifies the estimation problem and avoids observability issues, e.g., identical effects on unobservable or unmatched uncertainties~\cite{hovakimyan2010l1}.
%
% The third and final insight is that a system trained to adapt to a large variety of morphological and dynamical parameters is automatically robust to disturbances which are very difficult to model and to train on, \emph{e.g}, aerodynamics effects or swinging payloads.
%
% \todo{When the authors enumerate the three insights leveraged by their algorithm, the third comment about the algorithm being robust to disturbances within the convex hull of the training should be substantiated. }
%
The third and final insight is that a system can adapt to previously unseen disturbances as long as their effect on the platform dynamics is in the convex hull of the training disturbances (e.g., a motor losing efficiency has a similar effect to adding a payload below the motor).
%Therefore, future work should not spend effort on simulating all possible disturbances but on finding the minimal set of parameter changes to trigger a reaction.}
% we can train both the controller and the latent parameter identification module in simulation, where both the ground truth values and the state-action history are available. \antonio{The last insight is a bit less "insightful" than the others, maybe remove?}

To realize this, we follow an approach initially proposed for legged robots~\cite{kumar2021rma}. However, while that work performs online adaptation to terrains, we use their approach to adapt to a diverse set of quadcopters and perturbations. Specifically, we estimate a latent representation of the quadcopter's body from a history of sensor observations and actions, which conditions the behaviour of the controller.
%
%Such representation conditions the behaviour of the low-level controller.
% The space of quadcopter properties in which we train is extremely diverse (Table~\ref{tab:randomization}).
%
%\antonio{The flow here is okaish, but don't know how to improve it}.
%
The diversity of training parameters (Table~\ref{tab:randomization}) enables adaptation to sudden changes in environment conditions, e.g., a swing payload, for which it was not explicitly trained.
Our approach frees drone designers from the estimation and tuning process required any time something changes, or the risk that parameter changes unwittingly cause the control behaviour to significantly change, potentially endangering the system.
Furthermore, it naturally lends itself for use in off-the-shelf autopilot systems, allowing users who might otherwise lack the modelling abilities to control a custom vehicle, simply by plugging in the autopilot.

\begin{figure}
\centering
\includegraphics[width=\linewidth]{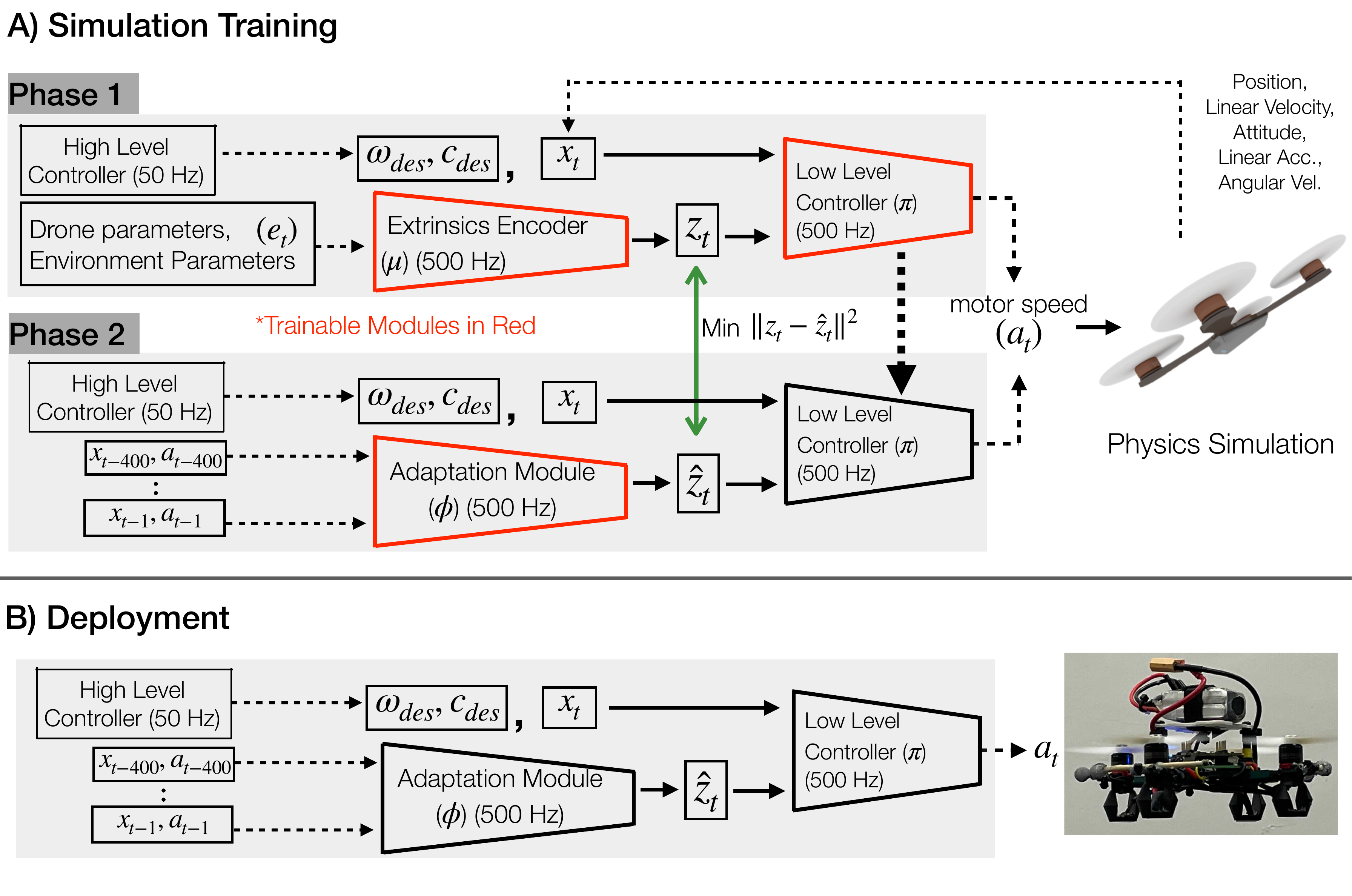}
\caption{The training (top) and the deployment architecture of our system (bottom). We train in two phases. In the first phase, we train a base policy $\pi$ which takes the current state $x_t$, and the intrinsics vector $z_t$ which is a compressed version of the environment parameters $e_t$ generated by the module $\mu$. Since we cannot deploy this policy in the real world because we do not observe the environment parameters $e_t$, we learn an adaptation module which takes the sensor history and action history, and directly predicts the intrinsics vector $z_t$. This is done in phase two in simulation using supervised learning. We can finally deploy the base policy $\pi$ which takes as input the current state $x_t$ and the intrinsics vector $\hat{z_t}$ predicted by the adaptation module $\phi$.}
\vspace{-2ex}
\label{fig:method}
\end{figure}

Our approach is related to existing methods in adaptive control.
However, our work shifts the meaning of adaptation to a different paradigm.
While adaptive control is generally concerned with estimating and counteracting disparities between observations and a \emph{reference} model, our approach does not have this notion. % of reference model.
This key difference enables adaptation to a much wider range of dynamics and disturbances.
In addition, it waives the engineering and tuning required by prior work when modifying the vehicle.
%
% Meta-learning is also related to the context of our problem in providing fast online adaptation.
% %
% Although they have been demonstrated on real quadcopters with impressive results on disturbance rejection~\cite{belkhale2021model,neuralfly}, they require real world learning samples to adapt.
% %
% When adapting to a much wider scale in our problem setting, this approach could be potentially dangerous for a dynamic system like a quadcopter, since failure to adapt quickly leads to catastrophic failure (a crash).
%
The closest related to our work are flight controllers in industrial systems like Beta-Flight~\cite{BetaFlight} and PX4~\cite{PX4}.
While they work well across a variety of vehicles, they still require significant in-flight tuning for each robot type and are generally tailored to human pilots.
\section{Related Work}
\label{sec:rel_work}

% \todo{Lit Review: Should include this work ‘Hardware Conditioned Policies for Multi-Robot Transfer Learning’ (I will refer to this paper as [1]) . This also uses a learned latent representation for the robot. }

% \todo{add LTF in lit rev} - no necessary with current Lit Review topic: adaptive control

% \todo{move related work ahead}

% \todo{Similar feedback to the discussion: Section II.B. does not provide a convincing argument as to why any of the existing learning-based methods (can solve the universal control problem). When comparing alternative methods, include the algorithms used in the experimental results to strengthen the argument as to why the universal controller has not been found yet.}

% \todo{Second of introduction should be incorporated into this section, but otherwise the writing is clear.}

% The design of high-performance adaptive controllers for aerospace systems has been a top priority for researchers and industry for more than 50 years.
% \todo{This isn't true, though? quadcopters may have been invented 100 years ago, but until we had small IMUs (maybe 20 years ago) no-one was really working on them, and certainly not a `top priority'.}
% \todo{We can maybe claim something like automatic control of aerospace systems...}
%
% While the overall goal remained the same over the decades, approaches greatly evolved.
%
% While each method has its advantages and limitations, they are mainly designed to handle model uncertainties and disturbances.
%
% We are interested in exploring a much wider range of adaptations, with substantial differences between platforms.

\subsection{Traditional Adaptive Control}
The design of high-performance adaptive controllers for aerospace systems has been a top priority for researchers and industry for more than 50 years.
One of the initial contributions in this space is the model reference adaptive controller (MRAC), an extension of the well-known MIT-rule~\cite{MAREELSMit}.
The empirical success of this method sparked great interest in the aerospace community, which led to the development of both practical tools and theoretical foundations~\cite{aastrom2013adaptive, lavretsky2013robust}.
From the many methods developed, one of the most popular is $\mathcal{L}_1$ adaptive control~\cite{cao2008design,hovakimyan2010l1}.
The main reason behind its success is the ability to provide rapid adaptation to model uncertainties and disturbances with theoretical guarantees under (possibly restrictive) assumptions.
Its high-level working principle consists of estimating the differences between the nominal (as predicted by the reference model) and observed state transitions.
Such differences are then compensated by allocating a control authority proportional to the disturbance, effectively driving the system to its reference behaviour.
Applications of $\mathcal{L}_1$ adaptive controllers for aerial vehicles span from multi-rotors to fixed wings~\cite{mallikarjunan2012l1, gregory2009l1}.

Mostly related to this work is the application of $\mathcal{L}_1$ adaptive control to quadcopters~\cite{schreier2012modeling}.
However, the performance of the classic $\mathcal{L}_1$ formulation degrades whenever the observed transitions differ greatly from the (usually linear) reference model, which can happen due to aerodynamic effects or large payloads.
Therefore, recent work has combined $\mathcal{L}_1$ adaptive control with nonlinear online optimization~\cite{hanover2021performance, pravitra2020, pereida2018adaptive}.
While these methods achieved impressive results, they still require explicit knowledge of (reference) system parameters, such as inertia, mass, and motor characteristics.
In addition, they generally require platform-specific tuning to get the desired behaviour. 
Other approaches to adaptive control on quadcopters include differential flatness~\cite{faessler2017differential} and nonlinear dynamic inversion~\cite{smeur2016adaptive}.
These methods have shown rapid adaptation to aerodynamic effects and model uncertainties.
However, they cannot cope with large variations in the quadcopter's dynamics, since the underlying assumptions on locally linear disturbances are generally not fulfilled.

\subsection{Learning-Based Adaptive Control}

Recent data-driven controllers have shown promising results for quadcopter stabilization~\cite{hwangbo2017control,koch2019reinforcement}, or waypoint tracking flight~\cite{song2021autonomous, kaufmann2022benchmark}.
As their classic counterparts, learning-based controllers also allow for adaptation to disturbances and model mismatches.
One possibility to do so consists of learning a model from the data and using the model to adapt the controller~\cite{lambert2019low,shi2019neural,belkhale2021model,torrente2021data}.
However, this has the limitation that the models are difficult to carefully identify due to under-actuation of the platform and sensing noise.
This motivated model-free methods that, like ours, learn an end-to-end adaptive policy~\cite{pi2021robust}.
Meta-learning has also been proposed to augment the performance of the model-based controller for fast online adaptation to wind~\cite{neuralfly} or suspended payloads~\cite{belkhale2021model}. 
These methods achieved impressive results in disturbance rejection.
However, they are still tailored to a specific platform type, and lack an explicit mechanism for adaptation to drastic changes in the drone's model and actuation. They also require relatively expensive real world learning samples. 
Our work aims to fill this gap and create a single control policy capable of flying vehicles with vastly varying physical characteristics and under large disturbances.
%
% However, we believe that meta-learning can complement our work on rapid adaptation by improving the policy in the real world during deployment. We will discuss this aspect in the future work and limitations section in the updated version.

% \begin{itemize}
%     %\item Model-based meta-reinforcement learning for flight with suspended payloads
%     %\item Neural lander: Stable drone landing control using learned dynamics
%     %\item Autonomous drone racing with deep reinforcement learning
%     %\item Low-level control of a quadcopter with deep model-based reinforcement learning
%     %\item Robust quadcopter control through reinforcement learning with disturbance compensation
%     %\item Low-level autonomous control and tracking of quadcopter using reinforcement learning
%     %\item Reinforcement Learning for UAV Attitude Control
%     %\item Neural-Fly enables rapid learning for agile flight in strong winds
%     %\item A Benchmark Comparison of Learned Control Policies for Agile quadcopter Flight
    
% \end{itemize}

\section{Learning an Adaptive Drone Controller}
\label{sec:method}

% \todo{Moving the definition intrinsics vector to the introductory paragraph before section A will clarify your high-level exposition before you dive into a more detailed description.}

% \todo{1. Algorithmic novelty As far as I understand, the proposed method is an application of RMA to the quadcopter control. Even though the authors mention that the proposed approach "re-purpose" the adaptation module of RMA to be suitable for the target task, as a reviewer, it was not clear to me the mentioned difference is significant. If the change is substantial, clearly spelling out the difference between the proposed method and RMA enhances the credibility of the manuscript.}

% \todo{- the definition of the 23-dimensional state is not included. }

% \todo{Also, when describing the network architecture, it would be clearer to separate better the description of the factor encoder and the adaptation module; and give more details of the latter for the sake of reproducibility.}

% \todo{- What is the window k used in the adaptation module (it seems the value of k is 400 according to page 4; 0.8 secs).}
We learn a single position controller to fly quadcopters to target positions and stay hovering. This controller takes a history of platform states and commands from a platform-independent high level controller as inputs, and outputs individual motor speed. Our resulting policy can control quadcopters with very different design and hardware characteristics, and is robust to disturbances unseen at training time, such as swing payloads and malfunctioning motors.
%In this work, we view both of these as adaptations to parameters intrinsic and extrinsic to drones, and we will learn a single adaptive policy capable of flying multiple different quadcopter drones under variations in payload and inertia. 
 To achieve this, we use RMA~\cite{kumar2021rma}. However, we re-purpose the adaption module, originally used to estimate parameters external to the robot, e.g. friction, to estimate the robot's intrinsic parameters (e.g. its mass, inertia, motor constant, etc.). In the following, we recapitulate the method of RMA for completeness. Our policy consists of a base policy $\pi$ and an adaptation module $\phi$. At time $t$, the base policy $\pi$ takes the current state $x_t \in \mathbb{R}^{23}$ and an intrinsics vector $z_t \in \mathbb{R}^6$ as input and outputs the target motor speeds $a_t \in \mathbb{R}^4$ for all individual motors. The current state $x_t$ includes the vehicle's position, velocity, rotation matrix, mass-normalized thrust, angular velocity, commanded total thrust and commanded angular velocity. The intrinsics vector $z_t$ is a low dimensional encoding of the environment parameters $e_t$ containing drone parameters, payload, etc. It allows the base policy to adapt to variations in drone parameters, payloads, and disturbances such as external force or torque. Since we cannot directly measure $z_t$ in the real world, we instead estimate it via the adaptation module $\phi$, which uses the discrepancy between the commanded actions and the measured sensor readings from the latest $k$ steps to estimate it online during deployment. More concretely, 
\begin{align}
    \hat{z_t} &=  \phi\big(x_{t-k:t-1}, a_{t-k:t-1}\big) \\
    a_t &= \pi(x_t, \hat{z_t}) \label{eq:pi}
\end{align}

\subsection{Base Policy}
We learn the base policy $\pi$ in simulation using model-free Reinforcement Learning (RL). During training, the base policy takes the current state $x_t$ and the ground-truth intrinsics vector $z_t$ to output control commands $a_t$. We use the environmental factor encoder $\mu$ to compress $e_t$ to $z_t$. This gives us:
\begin{align}
        z_t &= \mu(e_t) \\  
        a_t &= \pi(x_t, z_t)
\end{align}

% \todo{remove previous action from obs}

We implement $\mu$ and $\pi$ as Multi-layer perceptrons (MLP) and jointly train the base policy $\pi$ and the environmental factor encoder $\mu$ end-to-end using model-free RL. RL maximizes the following expected return of the policy $\pi$: 
\begin{align}
    J(\pi) = \mathbb{E}_{\tau \sim p(\tau|\pi)}\Bigg[\sum_{t=0}^{T-1}\gamma^t r_t\Bigg],
\end{align}
where $\tau = \{(x_0, a_0, r_0), (x_1, a_1, r_1) . . .\}$ is the trajectory of the agent when executing the policy $\pi$, and $p(\tau|\pi)$ represents the likelihood of the trajectory under $\pi$.

\paragraph{RL Reward} The following reward function encourages the agent to hover at a goal position and penalizes crashes and oscillating motions. Let us denote the acceleration in the body-fixed thrust axis, i.e. the mass-normalized thrust as $c$, the angular velocity as $\bm{\omega}$, all in the quadcopter's body frame. The commanded mass-normalized thrust is defined as $c_{des}$, commanded angular velocity as $\bm{\omega}_{des}$, all given by the high-level controller.
We additionally define the motor speeds as $\bm{m}$ and the simulation step in the training as $\delta t$. The reward at time $t$ is defined as the sum of the following quantities:
\begin{enumerate}
    \item Angular Velocity Tracking Deviation Penalty: \\$-\| \bm{\omega}^{t} - \bm{\omega}_{des}^{t} \|$
    \item Mass-normalized Thrust Tracking Deviation Penalty: $-\| {c}^{t} - {c}_{des}^{t} \|$
    \item Output Command Oscillation Penalty: $-\| \bm{m}^{t} - \bm{m}^{t-1} \|$
    \item Survival Reward: $\delta t$
\end{enumerate}
The scaling factor of each reward term is $0.01$, $0.02$, $0.0002$, $1$ respectively.

\begin{table}[t]
\caption{\label{tab:randomization} Ranges of the drone and environmental parameters. Parameters without units labelled are dimensionless quantities.}
\setlength{\tabcolsep}{5.5pt}
\begin{center}
\begin{tabular}{lcc}
\toprule
 \textbf{Parameters} & \textbf{Training Range} & \textbf{Testing Range} \\
 \midrule
 Mass (kg) & [0.142, 0.950]  & [0.114, 1.140]  \\
 Arm length (m) & [0.046, 0.200] & [0.037, 0.240] \\
 Mass moment of inertia  & \multirow{2}{*}{ [7.42e-5 , 5.60e-3]}&\multirow{2}{*}{ [5.94e-5 , 6.72e-3]} \\
%  [7.42e-5 , 5.60e-3 ] & [5.94e-5 , 6.72e-3 ]   \\
 around $x$, $y$ (kg$\cdot$m$^2$)&&\\
  Mass moment of inertia  & \multirow{2}{*}{[1.20e-4, 8.80e-3]} &  \multirow{2}{*}{[9.60e-5, 1.06e-2]} \\
  around $z$ (kg$\cdot$m$^2$)&&\\
 Propeller constant $\kappa$ (m) & [0.0041, 0.0168] & [0.0033, 0.0201]\\
 Payload (\% of Mass)  & [10, 50] & [5, 60] \\
 Payload Location & \multirow{3}{*}{[-10, 10]}  & \multirow{3}{*}{[-10, 10]}  \\
 from Center of Mass&&\\
 (\% of Arm length)&&\\
 Motor Constant & [1.15e-7, 7.64e-6] & [9.16e-8, 9.17e-6]  \\
 Body drag coefficient & [0, 0.62]& [0, 0.74] \\
 Max. motor speed (rad/s) & [707, 4895] & [566, 5874] \\
\bottomrule
\end{tabular}
% \marksez{$\kappa$ has units too, if it is "propellerTorqueFromThrust" from our lab sim code, the units are [m]. If it is really `torque to thrust' as in the paper, its units are [1/m].}
\end{center}
\end{table}

% \paragraph{Drone and Environment Randomization} We randomize the drone parameters which includes mass, arm length, inertia matrix, $\kappa$ ( propeller torque to thrust ratio constant), propeller thrust from speed squared constant, body drag coefficient, max motor speed and other external parameters like payload (Table~\ref{tab:randomization}). Note that we define a single vector $e_t$ which includes both the drone and the environment parameters and compress it into a single extrinsics vector $z_t$.  

\subsection{Adaptation Module}
During deployment, we do not have access to the vector $e_t$ and hence we cannot compute the intrinsics vector $z_t$. Instead, we will use the sensor and action history to directly estimate $z_t$ as proposed in \cite{kumar2021rma}. We call this module the adaptation module $\phi$. We can train this module in simulation using supervised learning because we have access to both the ground truth intrinsics $z_t$, and the sensor history and previous actions. We minimize the mean squared error loss $\lVert z - \hat{z} \rVert ^2$.  

\subsection{Deployment}
We directly deploy the base policy $\pi$ which uses the current state $x_t$ and the intrinsics vector $\hat{z_t}$ predicted by the adaptation module $\phi$. We do not calibrate or finetune our policy, and use the same policy without any modifications on the different drones under different payload and conditions.
\section{Experimental Setup}

\mypara{Simulation Environment} We use the Flightmare simulator~\cite{song2020flightmare} for training and testing our control policies. 
We implement the same high-level controller in \cite{mueller2018multicopter} to generates high-level commands at the level of body rates and collective thrust.
It is designed as a cascaded linear acceleration controller with desired acceleration mimicking a spring-mass-damper system with natural frequency 2rad/s and damping ratio 0.7. The desired acceleration is then converted to a desired total thrust and the desired thrust direction, and the body rates are computed from this as proportional to the attitude error angle, with a time constant of 0.2s.
% \marksez{The gains are a little hard to express compactly here. The rates are computed as (ignoring the nonlinear aspects) proportional to position, velocity, and angle error. I.e., the command has a "P" term, a "D" term, and a "double-D" term (as angle is like acceleration, to first order). 
% % Can perhaps describe it as follows:
% `We implement a custom high-level controller, that is designed as a cascaded linear acceleration controller (with desired acceleration mimicking a spring-mass-damper system with natural frequency 2rad/s and damping ratio 0.7. The desired acceleration is then converted to a desired total thrust, and desired thrust direction, and body rates are computed from this as proportional to the attitude error angle, with time constant 0.2s.'
% }
%
We define the task as hovering at a pre-defined location.
This high-level controller's inputs are the platform's state (position, rotation, angular, and linear velocities) and the goal location. 
The policy outputs individual motor speed commands, and we model the motors' response using a first-order system.
% We train the policy with model-free RL, using the PPO algorithm~\cite{schulman2017proximal}.
%
Each RL episode lasts for a maximum of 5s of simulated time, with early termination if the quadcopter height drops below 2cm.
The control frequency of the policy is $500$Hz, and the simulation step is 2ms.
We additionally implement an observation latency of 10ms.

\mypara{Quadcopter and Environment Randomization} All our training and testing ranges in simulation are listed in Table~\ref{tab:randomization}. Of these, $e_t$ includes mass, arm length, propeller torque to thrust ratio $\kappa$, motor constant, inertia ($\mathbb{R}^{3\times3}$), body drag coefficients ($\mathbb{R}^{3}$), maximum motor rotation and payload mass, which results in an 18 dimensional vector.
We randomize each of these parameters at the end of each episode. In addition, at a randomly sampled time during each episode, the parameters including mass, inertia, and the center of mass are also randomized. 
% \marksez{What does it mean to randomize at the end of an episode? Do we mean at the beginning of each episode? Isn't everything discarded at the end?}
%
The latter is used to mimic sudden variations in the quadcopter parameters due to a sudden disturbance caused by a payload or wind.

\mypara{Hardware Details} For all our real-world experiments we use two quadcopters, which differ in mass by a factor of 4.5, and in arm length by a factor of 2.9. 
The first one, which we name \emph{largequad} has a mass of 792g, a size of 16.6cm in arm length, a thrust-to-weight ratio of 3.50, a diagonal inertia matrix of [0.0047, 0.005, 0.0074]kg$\cdot$m$^2$ (as expressed in the z-up body-fixed frame), and a maximum motor speed of 943rad/s. The second one, \emph{miniquad}, has a mass of 177g, a size of 5.8cm in arm length, a thrust-to-weight ratio of 3.45, a diagonal inertia matrix of [92.7e-6, 92.7e-6, 158.57e-6]kg$\cdot$m$^2$, and a maximum motor speed of 3916rad/s.
A motion capture system running at $200$Hz provides estimates of the drone position and orientation, which is followed by a Kalman filter to reduce noise and estimate linear velocity.
An onboard rate gyroscope measures the angular rotation of the robot, which is low-pass filtered to reduce noise and remove outliers.
The deployed policy outputs motor speed commands, which are subsequently tracked by off-the-shelf electronic speed controllers.
We use as high-level a PD controller which takes as input the goal position and outputs the mass normalized collective trust and the body rates.

\mypara{Network Architecture and Training Procedure} The base policy is a 3-layer MLP with 256-dim hidden layers. This takes the drone state and the vector of intrinsics as input to produce motor speeds. The environment factor encoder is a 2-layer MLP with 128-dim hidden layers.  The policy and the value function share the same factor encoding layer. The adaptation module projects the latest 400 state-action pairs into a 128-dim representation, with the state-action history initialized with zeros. Then, a 3-layer 1-D CNN convolves the representation across time to capture its temporal correlation. The input channel number, output channel number, kernel size and stride of each CNN layer are [32, 32, 8, 4], [32, 32, 5, 1], [32, 32, 5, 1]. The flattened CNN output is linearly projected to estimate $z_t$. We train the base policy and the environment encoder using PPO~\cite{schulman2017proximal} for 100M steps. We use the reward outlined in Section~\ref{sec:method}. Policy training takes approximately 2 hours on an ordinary desktop machine with 1 GPU.
We finally train the adaptation module with supervised learning by rolling out the student policy. We train with the ADAM optimizer to minimize MSE loss. 
We run the optimization process for 10M steps, training on data collected over the last 1M steps. Training the adaptation module takes approximately 20 minutes.

\section{Results}

% \todo{I am unclear what the Thrust Err represent. Is the requested the Z acceleration?}

% \todo{Two more major changes which I believe are needed, more detailed investigation on unseen parameter performance, and more detail on the limitations of the work.}

% \todo{Paragraph breaks between the algorithms used in the experiment, the metrics used to compare the methods, and the final comparison would make the argument easier to read.}

% \begin{figure}[t]
%     \centering
%     \includegraphics[width=\linewidth]{img/Disturbances_2.png}
%     \caption{\textbf{Left}: Large and small quadcopters mounted with an inertia board. For the large quadcopter, we mount a wrench of 20.5cm and 140g. For the small quadcopter, a wrench of 14.5cm and 30g. \textbf{Right}: Large and small quadcopters with a rigid payload. For the former, we add a load of 180g ($\approx 25\%$ of its weight). For the latter, we add a load of 50g, which corresponds to 35.7\% of its weight. Despite such large disturbances unknown to the control policy, our approach can always stabilize the quadcopter. Videos at~\url{https://dz298.github.io/universal-drone-controller/}}
%     \label{fig:distur}
% \end{figure}

\begin{figure}[t]
    \centering
    \includegraphics[width=\linewidth]{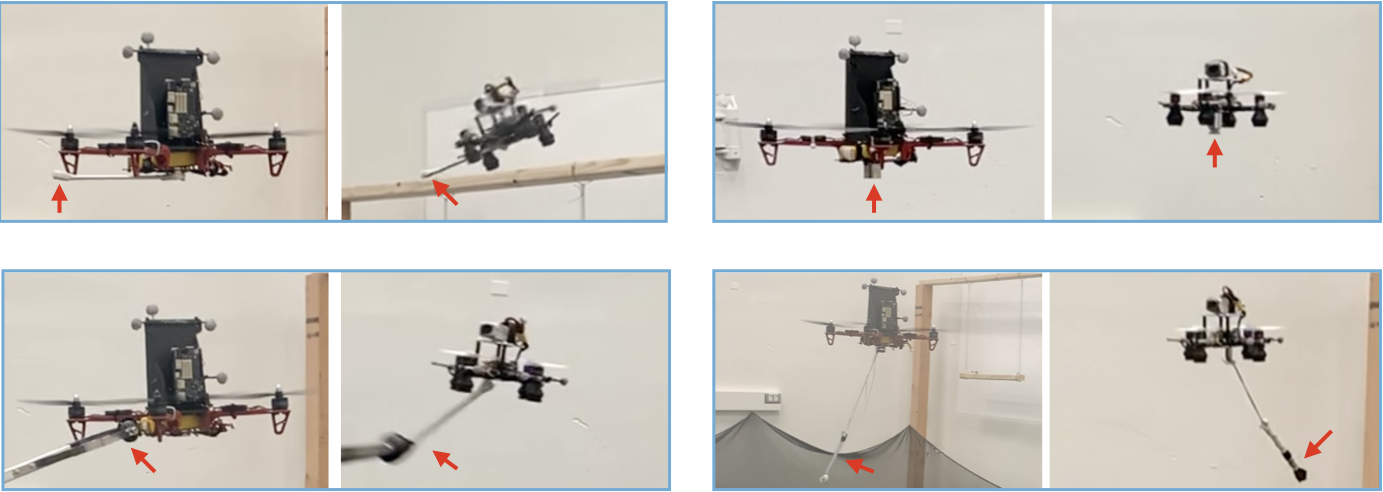}
    \caption{\textbf{Upper Left}: Large and small quadcopters mounted with an inertia board. For the large quadcopter, we mount a wrench of 20.5cm and 140g. For the small quadcopter, a wrench of 14.5cm and 30g. \textbf{Upper Right}: Large and small quadcopters with a rigid payload. For the former, we add a load of 180g ($\approx 25\%$ of its weight). For the latter, we add a load of 50g, which corresponds to 35.7\% of its weight. \textbf{Lower Left}: Large and small quadcopters with random pushes/pulls. \textbf{Lower Right}: Large and small quadcopters with a swing payload, which is the wrench in the inertia board test. Despite such large disturbances unknown to the control policy, our approach can always stabilize the quadcopter.}
    \vspace{-3ex}
    \label{fig:distur}
\end{figure}

\begin{table}[t]
\caption{\label{tab:rw_results} \textbf{Real-World  Results:} We compare the performance of our controller to two baselines: \emph{LTF} and a PID controller. The comparison is run on three tasks for large and small quadcopters. \textbf{Free Hover}: hover at the goal position without any disturbances. \textbf{Inertia Board}: hover at the goal position under an unknown mass and inertia disturbance. \textbf{Payload}: hover at the goal position under an unknown mass disturbance.  
% The \emph{LTF} learns a robust policy to all variations seen in the training by inputting an error integration into observations, while the PID controller has specifically been tuned to the two quadcopters with in-flight tests. We see that our approach significantly outperforms the two baseline in all metrics under all disturbances, particularly in inertia board case. Our method and the PID controller perform similarly on free hovering case with the PID controller sometimes outperforming our method. This is expected because the PID controller is specifically tuned for each quadcopter. 
Metrics are averaged over 5 experiments.}
\setlength{\tabcolsep}{3pt}
\def \arraystretch{1.4}
\centering
% \caption{Experimental results.}
% \label{tab:experimental_result.}
\begin{tabular}{lllllll}
\toprule
%   \multirow{2}{*}{\textbf{External Forces}} & Success & Height  & Ang. Vel. &  Thrust \\ &Rate& Err. (m) &Err. (rad/s)  & Err. (m/s$^2$) \\%& Learning Samples\\ 
 & \multirow{2}{*}{Vehicle}& \multirow{2}{*}{Method}& Height & Ang. Vel. & Thrust& Success\\&&&Err. (m)&Err. (rad/s)&Err. (m/s$^2$)& Rate\\
\midrule
\multirow{6}{*}{\begin{tabular}[c]{@{}l@{}}\textbf{Free} \\ \textbf{Hover}\end{tabular}} & \multirow{3}{*}{small} & PID & 0.06 & 0.51 & 0.57 &  100\\ %\cline{3-6} 
&  & LTF~\cite{LTF} & 0.09 & 0.78 & 0.78 & 80 \\
 &  & Ours & 0.05 & 0.30 & 0.30 & 100 \\ \cline{2-7} 
 & \multirow{3}{*}{large} & PID & 0.01 & 0.12 & 0.28 & 100 \\ %\cline{3-6} 
 &  & LTF~\cite{LTF} & 0.03 & 1.21 & 1.86 & 80 \\
 &  & Ours & 0.03 & 0.19 & 0.36 & 100 \\ \midrule

\multirow{6}{*}{\begin{tabular}[c]{@{}l@{}}\textbf{Inertial} \\ \textbf{Board}\end{tabular}} & \multirow{3}{*}{small} & PID & - & - & - &  0\\ %\cline{3-6} 
&  & LTF~\cite{LTF} & - & - & - & 0 \\
 &  & Ours & 0.08 & 1.14 & 1.09 & 100 \\ \cline{2-7} 
 & \multirow{3}{*}{large} & PID & 0.31 & 0.37 & 1.46 & 100 \\ %\cline{3-6} 
 &  & LTF~\cite{LTF} & - & - & - & 0 \\
 &  & Ours & 0.08 & 0.24 & 1.35 & 100 \\ \midrule

\multicolumn{1}{c}{\multirow{6}{*}{\textbf{Payload}}} & \multirow{3}{*}{small} & PID & 0.40 & 1.14 & 2.20 & 80 \\ %\cline{3-6} 
&  & LTF~\cite{LTF} & 0.10 & 0.98 & 2.14 & 100 \\
\multicolumn{1}{c}{} &  & Ours & 0.05 & 0.88 & 0.51 & 100 \\ \cline{2-7} 
\multicolumn{1}{c}{} & \multirow{3}{*}{large} & PID & 0.19 & 1.14 & 1.21 & 100 \\ %\cline{3-6} 
&  & LTF~\cite{LTF} & 0.06 & 1.01 & 4.2 & 100 \\
\multicolumn{1}{c}{} &  & Ours & 0.04 & 0.34 & 1.19 & 100 \\
\bottomrule
\end{tabular}
% \vspace{2ex}
\end{table}

\subsection{Real World Deployment}

% \todo{While the evaluation of the author’s method is thorough, PID serves as a  relatively brittle real-world comparison, especially when more capable methods (e.g. L1) exists in the literature. The reviewer recognizes that comparing against existing approaches (L1, etc.) in the real world requires extensive engineering effort, but when the authors claim they are developing the first ‘universal controller,’ such effort is warranted, especially if the main application of the method is an autopilot.  The comparison against LTF is reasonable, since it is also a model free approach of drones of a wider morphology (even from the ones tested in this paper!). However, since this paper has only tested x-framed quadcopters, [6] is also a valid real world comparison as well (for the same reason as simulation results above).}

We test our approach in the physical world and compare its performance to two baselines: \emph{LTF}, which is a learning-based robust controller trained with an error integration as one of inputs, aiming to control hybrid UAVs~\cite{LTF}; and a platform-specific PID controller.
The PID controller has access to the platform's mass and inertia, and it has been specifically tuned to the platform with in-flight tests.
In contrast, our approach has no knowledge whatsoever of the physical characteristics of the system and requires no calibration or real-world fine tuning.
We compare the task of stabilization to a predefined set point without any disturbances (free hover), or under an unknown mass and/or inertia disturbance (Figure~\ref{fig:distur}).

We compare the three approaches under four metrics: (i) the average height error to the goal point, and the average tracking error of the (ii) angular velocity and (iii) mass normalized thrust of the high-level controller's commands, and the success rate.
We define a failure if the human operator had to intervene to avoid the quadcopter from crashing.
The results of these experiments are reported in table~\ref{tab:rw_results}.

Our approach significantly outperforms both baselines in all metrics under all disturbances. In particular, our approach achieves a 100\% success rate when asymmetric disturbances are applied to the system, while the two baselines both experience at least one total failure on either of the two platforms. 
The PID baseline and our approach performs similarly in free hovering experiment, with the PID controller slightly outperforming ours on \emph{largequad}.  
The latter difference in performance is justified, since the PID controller is specifically tuned for each quadcopter, but ours does not have knowledge of the system dynamics and hardware. 
%
% Our approach significantly outperforms the PD baseline and the LTF baseline in all metrics. in terms of success rate, average height error, and tracking error on the applied thrust.
%
% The two methods perform similarly on tracking the angular velocity commands, with the PD slightly outperforming ours in some of the tests.
%
%However, our approach significantly outperforms the PD baseline in terms of average height error and tracking error on the applied thrust.
%
% The latter difference in performance is justified by the training procedure of our controller, which favours safety to tracking performance.
%
%Indeed, we strongly penalize the agent at training time for losing height or crashing into the ground.

\subsection{Simulation Results}

\begin{table}[h]
\caption{\label{tab:sim-metrics} \textbf{Simulation Testing Results:} We compare our method with four baselines on the task of stabilization: \emph{Robust}, \emph{SysID}, \emph{LTF}, and $\mathcal{L}_1$. We also list the results of our method in Phase I training, which has access to all ground-truth system parameters and can be regarded as the expert. The test ranges are defined in Table~\ref{tab:randomization}.
% The \emph{Robust} learns a single conservative flying method for all the variations seen during training; the \emph{SysID} baseline tries to predict the exact environment parameters which is both not difficult and not necessary to solve the task; the \emph{LTF} baseline learns a robust policy with an additional input of an error integration; and the $\mathcal{L}_1$ is a model-based adaptive controller which estimates and compensates for difference between the nominal and observed states. We observe that our method which is adaptive and estimates a low dimensional intrinsics outperforms all the baselines on the success rate and the average height tracking error. Our method also achieves comparable performance on the tracking of average angular velocity and normalized thrust with baselines best on these metrics. Our method's performance is most closest to the performance of the expert Phase I which has access to the ground truth system parameters. 
Metrics are averages over 100 experiments.
}
\centering
\begin{tabular*}{0.45\textwidth}{@{}lcccc@{}}
\toprule
   & Success & Height  & Ang. Vel. &  Thrust \\ &Rate& Err. (m) &Err. (rad/s)  & Err. (m/s$^2$) \\%& Learning Samples\\ 
\midrule
 Robust~\cite{peng2018sim,tobin2017domain} & 18\%& 0.44 & 1.32 & 1.93 \\ 
 SysID~\cite{SysID} & 38\%& 0.19 & 1.10 & 1.38\\
 LTF~\cite{LTF} & 59\% & 0.35 & 0.97 & 1.68 \\
 $\mathcal{L}_1$~\cite{hanover2021performance} & 59\% & 0.17 & 0.92 & 1.60 \\
 Ours & 66\% & 0.09 & 0.94 & 1.57 \\
 \midrule
 Expert (Phase I) & 69\% & 0.09 & 0.91 & 1.29 \\
\bottomrule
\end{tabular*}
% \begin{tabular}{cccccc}
% \toprule
%               & Robust & SysID & LTF     & L1   & Ours \\
%               \midrule
% success rate  & 18\%   & 38\%  & 59\% & 59\% & 66\% \\
% \midrule
% posx err      & 0.23   & 0.07  & 0.04    & 0.01 & 0.05 \\
% posy err      & 0.14   & 0.06  & 0.09    & 0.01 & 0.18 \\
% posz err      & 0.44   & 0.19  & 0.35    & 0.17 & 0.09 \\
% \midrule
% velx err      & 0.09   & 0.08  & 0.09    & 0.11 & 0.04 \\
% vely err      & 0.10   & 0.02  & 0.14    & 0.11 & 0.12 \\
% velz err      & 0.14   & 0.07  & 0.16    & 0.16 & 0.08 \\
% \midrule
% thrust err    & 1.93   & 1.38  & 1.68    & 1.60 & 1.57 \\
% ang vel z err & 1.65   & 0.89  & 0.72    & 0.21 & 0.78 \\
% ang vel x err & 1.15   & 1.35  & 1.09    & 1.28 & 0.93 \\
% ang vel y err & 1.16   & 1.06  & 1.10    & 1.25 & 1.09 \\
% \bottomrule
% \end{tabular}
\end{table}

\begin{table}[h]
\caption{\label{tab:sim-metrics-unseen-disturb} \textbf{Simulation Testing Results, Out-of-Distribution Disturbances}: We evaluate the performance of our method and all baselines on two types of disturbances unseen at training time. \textbf{External Forces}: We apply a random force of magnitude uniformly sampled between 0 and 50\% of the weight and with direction uniformly sampled on a cube. \textbf{Partially Failing Motors}: To simulate a motor losing efficiency, we multiply the output of a randomly sampled motor’s thrust force to a random number between 0 and 1. The duration of each disturbance is random between the entire length of the episode (on and off with 2\% probability at every time stamp).
}
\centering
\begin{tabular*}{0.45\textwidth}{@{}lcccc@{}}
\toprule
  \multirow{2}{*}{\textbf{External Forces}} & Success & Height  & Ang. Vel. &  Thrust \\ &Rate& Err. (m) &Err. (rad/s)  & Err. (m/s$^2$) \\%& Learning Samples\\ 
\midrule
 Robust~\cite{peng2018sim,tobin2017domain} & 5\%& 0.39 & 2.05 & 1.51 \\ 
 SysID~\cite{SysID} & 2\%& 0.22 & 1.38 & 1.23\\
 LTF~\cite{LTF} & 32\% & 0.23 & 1.53 & 0.99 \\
 $\mathcal{L}_1$~\cite{hanover2021performance} & 42\% & 0.30 & 1.46 & 1.04 \\
 Ours & 49\% & 0.09 & 1.40 & 0.85\\
 \midrule
 \multicolumn{2}{@{}l}{\textbf{Partially Failing Motors}} &&& \\
 Robust~\cite{peng2018sim,tobin2017domain} & 1\%& 0.53 & 1.94 & 1.59 \\ 
 SysID~\cite{SysID} & 14\%& 0.33 & 1.25 & 1.23\\
 LTF~\cite{LTF} & 21\% & 0.38 & 1.57 & 1.35 \\
 $\mathcal{L}_1$~\cite{hanover2021performance} & 33\% & 0.32 & 1.41 & 1.06 \\
 Ours & 38\% & 0.26 & 1.36 & 1.01\\
\bottomrule
\end{tabular*}
\end{table}

Finally, we compare our approach with a set of baselines in the simulation.
We select four baselines from prior work: \emph{Robust}, which consists of a policy trained without access to environment factors or body parameters~\cite{tobin2017domain, peng2018sim}; \emph{SysID}, which directly predicts the ground-truth parameters $e_t$~\cite{SysID} instead of the low-dimensional intrinsics vector; \emph{LTF}, which essentially is a robust policy with an error integration as additional inputs~\cite{LTF}; and $\mathcal{L}_1$, a model-based adaptive controller which estimates and compensates the difference between the nominal and observed states to achieve adaptive control~\cite{cao2008design,hovakimyan2010l1,hanover2021performance}. 
We keep the same architecture and hyperparameters as ours for all learning-based baselines. 
%
% We contacted the author of a state-of-the-art $\mathcal{L}_1$ adaptive controller ~\cite{hanover2021performance} to ensure the correctness of our implementation.\daisy{not a fan of this sentence, will remove}
%
Similarly to real-world experiments, we evaluate on the task of stabilization. The testing ranges are listed in Table~\ref{tab:randomization}. We rank the methods according to the success rate, the average height error, and the tracking performance.
At the beginning of each experiment, the quadcopter is spawned with a random orientation, a random position in x-y of [-1,1] and [0.5, 2.5] in z, and a random velocity on each axis of [-1,1].
The experiment is considered successful if the end height of the quadcopter is within 0.3m from the goal height.
The results of these experiments are reported in Table~\ref{tab:sim-metrics}. 

Given the very large amount of quadcopter variations, the \emph{Robust} baseline trained without access to environment parameters has the lowest success rate and largest tracking error. This is because it is forced to learn a single conservative controller which can fly all quadcopters under varying disturbances.
Indeed, it either crashes to the ground or flies outside the flying region.
The baseline \emph{LTF} is \emph{Robust} with an additional observation of error integration. This additional input helps it achieve higher success rate. However, similar to the \emph{Robust} baseline, \emph{LTF} fails in tracking the goal states. 
On the contrary, with estimation of the environment parameters, the flight performance strongly increases.
\begin{figure}[t]
    \centering
    \includegraphics[width=\columnwidth]{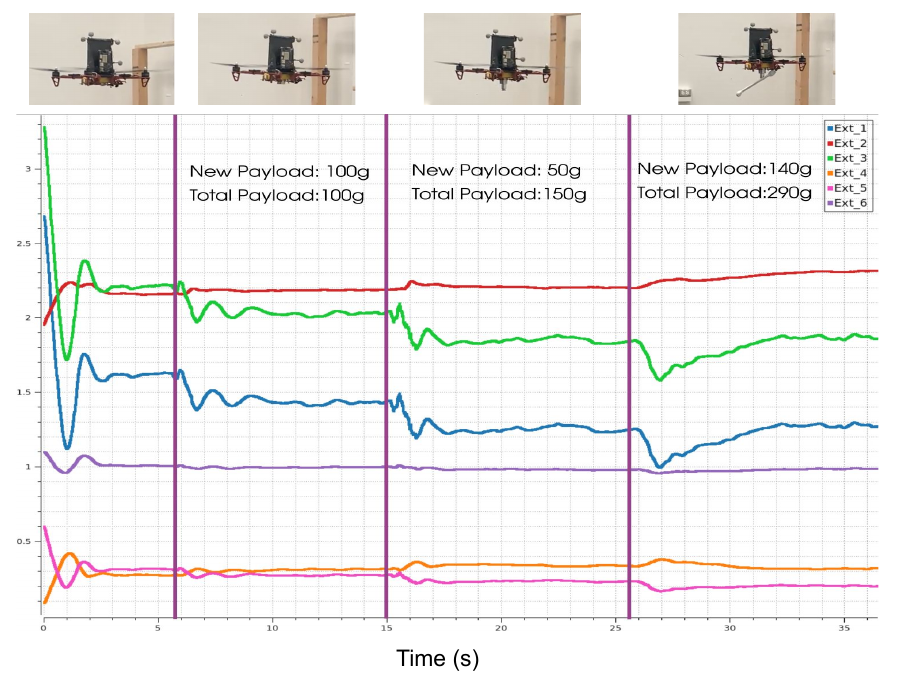}
    \caption{We analyze the change in behaviour of our policy as we incrementally add total 290g payloads to our \emph{largequad}. We plot all 6 components of the intrinsics vector $\hat{z_t}$ predicted by the adaptation module. We see that changes in intrinsics are strongly correlated with disturbances applied to the quadcopter, indicating that the added payloads have been detected by the adaptation module. When the payload is added, the quadcopter first sinks and then recovers to the normal motion. The plotted components of the intrinsics vector change in response to the disturbance, from which we know the adaptation period takes around 2s.}
    \label{fig:intrinsics_analysis}
\end{figure}
Still, explicitly regressing the environment vector $e_t$, instead of its low-dimensional representation $z_t$, yields the success rate since it is trying to solve a harder estimation problem which is not necessary to solve the task at hand.
Since the tracking errors are computed only for successful runs, the \emph{SysID} baseline achieves a slightly lower tracking error in one of the metrics.
$\mathcal{L}_1$ is the baseline with the strongest performance in our simulation experiments. However, its strong performance relies on the explicit knowledge of a reference model which we choose as the median value of all parameters in the testing range in Table~\ref{tab:randomization}, while our method with other baselines has no prior knowledge of the model. 
The performance gap between our method and $\mathcal{L}_1$ shows that the adaptive control across platforms is hard to achieve by estimating and compensating large variance from a reference model. 
$\mathcal{L}_1$ is not implemented in our real-world experiments as a baseline because it is highly sensitive to latency and requires significant engineering work. In contrast, our approach can handle latency very well by simulating it during training.

We also evaluate the task of stabilization on held-out environmental disturbances, as exemplified by random external forces and partial failure of a motor. The results of these experiments are reported in Table~\ref{tab:sim-metrics-unseen-disturb}. Our method outperforms all baselines in both out-of-distribution disturbances cases. Compared to the strongest baseline $\mathcal{L}_1$, our method has a greater success rate, and it reduces the average height error by up to 67.7\% and the average command tracking error by up to 22.4\%. 

% %
% Given the very large amount of quadcopter variations, the policy trained without access to environment parameters fails to achieve reasonable performance. This is because it is forced to learn a single conservative flying which can fly all quadcopters under varying disturbances.
% %
% Indeed, it either crashes to the ground or flies outside of the flying region.
% %
% Conversely, with access to environment parameters, the flight performance strongly increases.
% %
% Still, explicitly regressing the environment vector $e_t$, instead of its low-dimensional representation $z_t$ scarifies the success rate since it is trying to solve a harder estimation problem which is not necessary to solve the task at hand.
% %
% Since the tracking errors are computed only for successful runs, the \emph{SysID} baseline achieves a slightly lower tracking error on one of the metrics.
% %
% Having to solve additional scenarios which are potentially more challenging, our method accumulates a higher error, but still manages to keep the quadcopter from crashing.

% Our experiments show that whenever the policy  the importance of online adaptation.
% %
% We create a large 

\subsection{Adaptation Analysis}
We analyze the estimated intrinsics vector $\hat{z_t}$ for adaptation on incremental payloads. We incrementally add payloads in total of 290g to the \emph{largequad} in hovering. The quadcopter adapts successfully to payloads and stabilizes itself at the target position. 
We plot all components of the estimated intrinsics vector from the adaptation module during the experiment in Figure~\ref{fig:intrinsics_analysis}. 
We find that whenever a payload is applied, there is a change in all intrinsics components. 
%
% Then the quadcopter recovers from the disturbance and all components of the intrinsics vector converges to different values as they are before the payload is applied. 
% It shows that even after adaptation, the adaptation module still detects the existence of payloads. 
%
From the change of the plotted components of intrinsics vector in response to the disturbance, we see that it takes around 2s for our controller to detect the disturbance, estimate the intrinsics vector, adapt to the disturbance, and remain hovering at the target location. 
The adaptation period of our controller is much faster than that of methods with online estimation of ground truth model parameters such as~\cite{wuest2019online}, where it takes 10s to 15s in-flight to obtain an accurate model estimate.

\section{Conclusion}
In this work, we show how a single policy can control quadcopters of totally different size, mass, and actuation. 
We successfully transfer a method initially developed for legged locomotion in adapting terrains to quadcopters in adapting a diverse set of quadcopter bodies and disturbances.
Without any additional tuning or modification, the single policy trained only in simulation can be deployed zero-shot to quadcopters of very different design and hardware characteristics while showing rapid adaptation to unknown disturbances at the same time.
However, our current controller is a quasi-static position controller which fails on tracking aggressive flights. 
Future work should focus on improving its ability to track arbitrary trajectories to achieve a universal quadcopter controller. 
% Adaptive control has achieved great success in adaptation to model uncertainties and disturbances, but our work offers a fresh view on the meaning of adaptation by completely bypassing the notion of a \emph{reference} model that most adaptive control methods use. 
%
%while additionally showing rapid adaptation to unknown payloads during deployment. It differs from classic adaptative control in that it does not compensates for disparities between observations and the referenced model.
%
% This feature enables our method to adapt to a much wider range of robots and disturbances.
%
%
% One limitation of our approach is that it relies on simulation to improve, necessitating the need to replicate and train on situations in the simulation in which it might fail to work in the real world. A more scalable lifelong solution to the problem would be to continuously learn in the real world with the data collected during deployment in the real world. 
%
% \propose{Future work will aim to incorporate learning with real world data into the method. }
%
%In addition, we noticed a trade-off between the tracking performance and the probability of crashing. In this work, we favoured safety to tracking error. 
%and the  the policy presented in this paper is trained to take avoiding crashes at the time of adaptation as the priority out of other goals including tracking high-level commands. 
%However, future work will aim to relax this trade-off. 
%However, in the near future, we will have to improve the tracking performance and ensure safety.  

\section{Acknowledgement}
This work was supported by the DARPA Machine Common Sense program, Hong Kong Center for Logistics Robotics, and the Graduate Division Block Grant of the Dept. of Mechanical Engineering, UC Berkeley.
 The experimental testbed at the HiPeRLab is the result of contributions of many people, a full list of which can be found at \url{hiperlab.berkeley.edu/members/}. This work was also partially supported by the ONR MURI award N00014-21-1-2801.

\addtolength{\textheight}{-4.5cm}   % This command serves to balance the column lengths

%===============================================================================

% The maximum paper length is 8 pages excluding references and acknowledgements, and 10 pages including references and acknowledgements

%\clearpage
% The acknowledgments are automatically included only in the final and preprint versions of the paper.
% \acknowledgments{}

%===============================================================================

% \bibliography{references.bib}  % .bib

\bibliographystyle{IEEEtran}
\bibliography{references}

\begin{thebibliography}{10}
\providecommand{\url}[1]{#1}
\csname url@samestyle\endcsname
\providecommand{\newblock}{\relax}
\providecommand{\bibinfo}[2]{#2}
\providecommand{\BIBentrySTDinterwordspacing}{\spaceskip=0pt\relax}
\providecommand{\BIBentryALTinterwordstretchfactor}{4}
\providecommand{\BIBentryALTinterwordspacing}{\spaceskip=\fontdimen2\font plus
\BIBentryALTinterwordstretchfactor\fontdimen3\font minus
  \fontdimen4\font\relax}
\providecommand{\BIBforeignlanguage}[2]{{%
\expandafter\ifx\csname l@#1\endcsname\relax
\typeout{** WARNING: IEEEtran.bst: No hyphenation pattern has been}%
\typeout{** loaded for the language `#1'. Using the pattern for}%
\typeout{** the default language instead.}%
\else
\language=\csname l@#1\endcsname
\fi
#2}}
\providecommand{\BIBdecl}{\relax}
\BIBdecl

\bibitem{svacha2020imu}
J.~Svacha, J.~Paulos, G.~Loianno, and V.~Kumar, ``Imu-based inertia estimation
  for a quadrotor using newton-euler dynamics,'' \emph{IEEE Robotics and
  Automation Letters}, vol.~5, no.~3, pp. 3861--3867, 2020.

\bibitem{wuest2019online}
V.~W{\"u}est, V.~Kumar, and G.~Loianno, ``Online estimation of geometric and
  inertia parameters for multirotor aerial vehicles,'' in \emph{2019
  International Conference on Robotics and Automation (ICRA)}.\hskip 1em plus
  0.5em minus 0.4em\relax IEEE, 2019, pp. 1884--1890.

\bibitem{forgione2021continuous}
M.~Forgione and D.~Piga, ``Continuous-time system identification with neural
  networks: Model structures and fitting criteria,'' \emph{European Journal of
  Control}, vol.~59, pp. 69--81, 2021.

\bibitem{hovakimyan2010l1}
N.~Hovakimyan and C.~Cao, \emph{L1 adaptive control theory: Guaranteed
  robustness with fast adaptation}.\hskip 1em plus 0.5em minus 0.4em\relax
  SIAM, 2010.

\bibitem{kumar2021rma}
A.~Kumar, Z.~Fu, D.~Pathak, and J.~Malik, ``Rma: Rapid motor adaptation for
  legged robots,'' \emph{RSS: Robotics Science and Systems}, 2021.

\bibitem{BetaFlight}
``Betaflight: Open source flight controller firmware,''
  \url{https://betaflight.com/}.

\bibitem{PX4}
``Open source autopilot for drones: Px4 autopilot,'' \url{https://px4.io/}.

\bibitem{MAREELSMit}
I.~M. Mareels, B.~D. Anderson, R.~R. Bitmead, M.~Bodson, and S.~S. Sastry,
  ``Revisiting the mit rule for adaptive control,'' \emph{IFAC Proceedings
  Volumes}, vol.~20, no.~2, pp. 161--166, 1987.

\bibitem{aastrom2013adaptive}
K.~J. {\AA}str{\"o}m and B.~Wittenmark, \emph{Adaptive control}.\hskip 1em plus
  0.5em minus 0.4em\relax Courier Corporation, 2013.

\bibitem{lavretsky2013robust}
E.~Lavretsky and K.~A. Wise, ``Robust adaptive control,'' in \emph{Robust and
  adaptive control}.\hskip 1em plus 0.5em minus 0.4em\relax Springer, 2013, pp.
  317--353.

\bibitem{cao2008design}
C.~Cao and N.~Hovakimyan, ``Design and analysis of a novel l1 adaptive control
  architecture with guaranteed transient performance,'' \emph{IEEE Transactions
  on Automatic Control}, vol.~53, no.~2, pp. 586--591, 2008.

\bibitem{mallikarjunan2012l1}
S.~Mallikarjunan, B.~Nesbitt, E.~Kharisov, E.~Xargay, N.~Hovakimyan, and
  C.~Cao, ``L1 adaptive controller for attitude control of multirotors,'' in
  \emph{AIAA guidance, navigation, and control conference}, 2012, p. 4831.

\bibitem{gregory2009l1}
I.~Gregory, C.~Cao, E.~Xargay, N.~Hovakimyan, and X.~Zou, ``L1 adaptive control
  design for nasa airstar flight test vehicle,'' in \emph{AIAA guidance,
  navigation, and control conference}, 2009, p. 5738.

\bibitem{schreier2012modeling}
M.~Schreier, ``Modeling and adaptive control of a quadrotor,'' in \emph{2012
  IEEE international conference on mechatronics and automation}.\hskip 1em plus
  0.5em minus 0.4em\relax IEEE, 2012, pp. 383--390.

\bibitem{hanover2021performance}
D.~Hanover, P.~Foehn, S.~Sun, E.~Kaufmann, and D.~Scaramuzza, ``Performance,
  precision, and payloads: Adaptive nonlinear mpc for quadrotors,'' \emph{IEEE
  Robotics and Automation Letters}, vol.~7, no.~2, pp. 690--697, 2021.

\bibitem{pravitra2020}
J.~Pravitra, K.~A. Ackerman, C.~Cao, N.~Hovakimyan, and E.~A. Theodorou,
  ``L1-adaptive mppi architecture for robust and agile control of
  multirotors,'' in \emph{2020 IEEE/RSJ International Conference on Intelligent
  Robots and Systems (IROS)}.\hskip 1em plus 0.5em minus 0.4em\relax IEEE,
  2020, pp. 7661--7666.

\bibitem{pereida2018adaptive}
K.~Pereida and A.~P. Schoellig, ``Adaptive model predictive control for
  high-accuracy trajectory tracking in changing conditions,'' in \emph{2018
  IEEE/RSJ International Conference on Intelligent Robots and Systems
  (IROS)}.\hskip 1em plus 0.5em minus 0.4em\relax IEEE, 2018, pp. 7831--7837.

\bibitem{faessler2017differential}
M.~Faessler, A.~Franchi, and D.~Scaramuzza, ``Differential flatness of
  quadrotor dynamics subject to rotor drag for accurate tracking of high-speed
  trajectories,'' \emph{IEEE Robotics and Automation Letters}, vol.~3, no.~2,
  pp. 620--626, 2017.

\bibitem{smeur2016adaptive}
E.~J. Smeur, Q.~Chu, and G.~C. De~Croon, ``Adaptive incremental nonlinear
  dynamic inversion for attitude control of micro air vehicles,'' \emph{Journal
  of Guidance, Control, and Dynamics}, vol.~39, no.~3, pp. 450--461, 2016.

\bibitem{hwangbo2017control}
J.~Hwangbo, I.~Sa, R.~Siegwart, and M.~Hutter, ``Control of a quadrotor with
  reinforcement learning,'' \emph{IEEE Robotics and Automation Letters},
  vol.~2, no.~4, pp. 2096--2103, 2017.

\bibitem{koch2019reinforcement}
W.~Koch, R.~Mancuso, R.~West, and A.~Bestavros, ``Reinforcement learning for
  uav attitude control,'' \emph{ACM Transactions on Cyber-Physical Systems},
  vol.~3, no.~2, pp. 1--21, 2019.

\bibitem{song2021autonomous}
Y.~Song, M.~Steinweg, E.~Kaufmann, and D.~Scaramuzza, ``Autonomous drone racing
  with deep reinforcement learning,'' in \emph{2021 IEEE/RSJ International
  Conference on Intelligent Robots and Systems (IROS)}.\hskip 1em plus 0.5em
  minus 0.4em\relax IEEE, 2021, pp. 1205--1212.

\bibitem{kaufmann2022benchmark}
E.~Kaufmann, L.~Bauersfeld, and D.~Scaramuzza, ``A benchmark comparison of
  learned control policies for agile quadrotor flight,'' \emph{arXiv preprint
  arXiv:2202.10796}, 2022.

\bibitem{lambert2019low}
N.~O. Lambert, D.~S. Drew, J.~Yaconelli, S.~Levine, R.~Calandra, and K.~S.
  Pister, ``Low-level control of a quadrotor with deep model-based
  reinforcement learning,'' \emph{IEEE Robotics and Automation Letters},
  vol.~4, no.~4, pp. 4224--4230, 2019.

\bibitem{shi2019neural}
G.~Shi, X.~Shi, M.~O’Connell, R.~Yu, K.~Azizzadenesheli, A.~Anandkumar,
  Y.~Yue, and S.-J. Chung, ``Neural lander: Stable drone landing control using
  learned dynamics,'' in \emph{2019 International Conference on Robotics and
  Automation (ICRA)}.\hskip 1em plus 0.5em minus 0.4em\relax IEEE, 2019, pp.
  9784--9790.

\bibitem{belkhale2021model}
S.~Belkhale, R.~Li, G.~Kahn, R.~McAllister, R.~Calandra, and S.~Levine,
  ``Model-based meta-reinforcement learning for flight with suspended
  payloads,'' \emph{IEEE Robotics and Automation Letters}, vol.~6, no.~2, pp.
  1471--1478, 2021.

\bibitem{torrente2021data}
G.~Torrente, E.~Kaufmann, P.~F{\"o}hn, and D.~Scaramuzza, ``Data-driven mpc for
  quadrotors,'' \emph{IEEE Robotics and Automation Letters}, vol.~6, no.~2, pp.
  3769--3776, 2021.

\bibitem{pi2021robust}
C.-H. Pi, W.-Y. Ye, and S.~Cheng, ``Robust quadrotor control through
  reinforcement learning with disturbance compensation,'' \emph{Applied
  Sciences}, vol.~11, no.~7, p. 3257, 2021.

\bibitem{neuralfly}
M.~O’Connell, G.~Shi, X.~Shi, K.~Azizzadenesheli, A.~Anandkumar, Y.~Yue, and
  S.-J. Chung, ``Neural-fly enables rapid learning for agile flight in strong
  winds,'' \emph{Science Robotics}, vol.~7, no.~66, p. eabm6597, 2022.

\bibitem{song2020flightmare}
Y.~Song, S.~Naji, E.~Kaufmann, A.~Loquercio, and D.~Scaramuzza, ``Flightmare: A
  flexible quadrotor simulator,'' in \emph{Proceedings of the 2020 Conference
  on Robot Learning}, 2021, pp. 1147--1157.

\bibitem{mueller2018multicopter}
M.~W. Mueller, ``Multicopter attitude control for recovery from large
  disturbances,'' \emph{arXiv preprint arXiv:1802.09143}, 2018.

\bibitem{schulman2017proximal}
J.~Schulman, F.~Wolski, P.~Dhariwal, A.~Radford, and O.~Klimov, ``Proximal
  policy optimization algorithms,'' \emph{arXiv preprint arXiv:1707.06347},
  2017.

\bibitem{LTF}
J.~Xu, T.~Du, M.~Foshey, B.~Li, B.~Zhu, A.~Schulz, and W.~Matusik, ``Learning
  to fly: computational controller design for hybrid uavs with reinforcement
  learning,'' \emph{ACM Transactions on Graphics (TOG)}, vol.~38, no.~4, pp.
  1--12, 2019.

\bibitem{peng2018sim}
X.~B. Peng, M.~Andrychowicz, W.~Zaremba, and P.~Abbeel, ``Sim-to-real transfer
  of robotic control with dynamics randomization,'' in \emph{2018 IEEE
  international conference on robotics and automation (ICRA)}.\hskip 1em plus
  0.5em minus 0.4em\relax IEEE, 2018, pp. 3803--3810.

\bibitem{tobin2017domain}
J.~Tobin, R.~Fong, A.~Ray, J.~Schneider, W.~Zaremba, and P.~Abbeel, ``Domain
  randomization for transferring deep neural networks from simulation to the
  real world,'' in \emph{2017 IEEE/RSJ international conference on intelligent
  robots and systems (IROS)}.\hskip 1em plus 0.5em minus 0.4em\relax IEEE,
  2017, pp. 23--30.

\bibitem{SysID}
\BIBentryALTinterwordspacing
W.~Yu, J.~Tan, C.~K. Liu, and G.~Turk, ``Preparing for the unknown: Learning a
  universal policy with online system identification,'' in \emph{Robotics:
  Science and Systems XIII, Massachusetts Institute of Technology, Cambridge,
  Massachusetts, USA, July 12-16, 2017}, N.~M. Amato, S.~S. Srinivasa,
  N.~Ayanian, and S.~Kuindersma, Eds., 2017. [Online]. Available:
  \url{http://www.roboticsproceedings.org/rss13/p48.html}
\BIBentrySTDinterwordspacing

\end{thebibliography}

\end{document}